\documentclass[conference]{IEEEtran}
\IEEEoverridecommandlockouts
\usepackage{cite}
\usepackage{amsmath,amssymb,amsfonts}
\usepackage{algorithm2e}
\RestyleAlgo{ruled}
\usepackage{graphicx}
\usepackage{textcomp}
\usepackage{xcolor}
\usepackage{xcolor,colortbl}
\def\BibTeX{{\rm B\kern-.05em{\sc i\kern-.025em b}\kern-.08em
    T\kern-.1667em\lower.7ex\hbox{E}\kern-.125emX}}
\begin{document}

\title{Parameter Optimization of LLC-Converter with multiple operation points using \\ Reinforcement Learning}

\author{
\IEEEauthorblockN{Georg Kruse, Dominik Happel, Stefan Ditze, Stefan Ehrlich, Andreas Rosskopf}

\IEEEauthorblockA{\textit{Fraunhofer Institute for Integrated Systems and Device Technology IISB} \\
Erlangen, Germany \\
georg.kruse@iisb.fraunhofer.de}
}

\maketitle
\definecolor{light-gray}{gray}{0.95}
\begin{abstract}
The optimization of electrical circuits is a difficult and time-consuming process performed by experts, but also increasingly by sophisticated algorithms. In this paper, a reinforcement learning (RL) approach is adapted to optimize a LLC converter at multiple operation points corresponding to different output powers at high converter efficiency at different switching frequencies. During a training period, the RL agent learns a problem specific optimization policy enabling optimizations for any objective and boundary condition within a pre-defined range. The results show, that the trained RL agent is able to solve new optimization problems based on LLC converter simulations using Fundamental
Harmonic Approximation (FHA) within 50 tuning steps for two operation points with power efficiencies greater than 90\%. Therefore, this AI technique provides the potential to augment expert-driven design processes with data-driven strategy extraction in the field of power electronics and beyond.
\end{abstract}


\section{Introduction}

During the last decade, reinforcement learning (RL) has outperformed humans in strategy games like Chess, Go \cite{alphazero} or Starcraft \cite{starcraft} and also demonstrated first impressive results in engineering and related fields.  In \cite{alphafold} new protein structures were designed with this method, while in \cite{muzero} and \cite{pcb} the positioning of plasma in a magnetic field and the component routing on a PCB were optimized. In all cases, this special machine learning approach learns a policy of a so called RL agent, which solves engineering problems based on several thousands of iterations and interactions in a simulation environment. 

In this paper we first demonstrate the potential of a Proximal Policy Optimization (PPO) \cite{ppo} RL agent for the optimization of an electric circuit at multiple operation points. This approach outperforms current multi-objective optimization methods using genetic algorithms in terms of the number of evaluations \cite{rosskopf} and state-of-the-art RL agents in the domain of power electronics in the amount and range of the input parameters as well as objectives \cite{distributed}. 

The described use case targets resonant converters using LLC resonant circuits, which are important in the design of galvanically isolated switch-mode power supplies and inductive power transfer systems in industrial, automotive, and avionic applications. The LLC resonant converter depicted in Fig. 1 uses sinusoidal oscillations between the inductances $L_r$, $L_m$ and the resonant capacitance $C_r$ offering the advantages of zero-voltage switching in the applied semiconductors, size reduction of passive components by selecting high switching frequency, and advantageous utilization of parasitic elements (stray inductances). 

\begin{figure}[ht]
\centerline{\includegraphics[width=0.4 \textwidth]{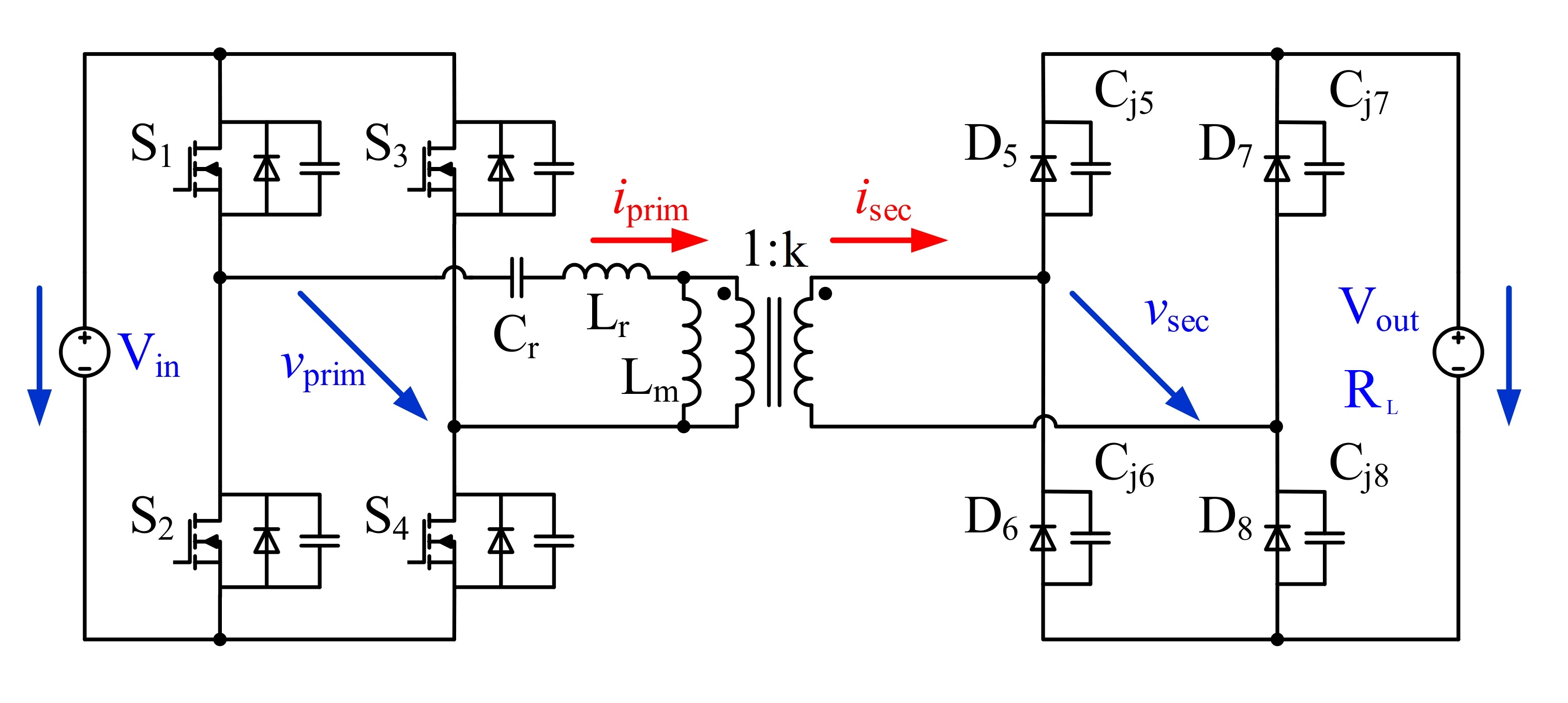}}
\caption{Topology of the LLC resonant converter}
\label{fig1}
\end{figure}

\section{Problem Statement}

LLC circuit simulations using accurate models of the semiconductor devices are always very CPU-intensive. To provide engineers with a first intuition of a good starting point for a detailed design process, we will use a fast, static approximation of the electrical behavior of the LLC converter based on analytic equations using Fundamental Harmonic Approximation (FHA) \cite{FHA}. 

This approximation algorithm takes input voltage $V_{in}$, the equivalent load resistance $R_L$, switching frequency $f_i$, inductances $L_r$ and $L_m$, resonant capacitance $C_r$ and coupling factor $k$ as input and computes the output power and the corresponding efficiency at the load resistance $R_L$.
In applications such as for electric vehicles charging, the system needs to work efficiently at multiple operation points corresponding to a set of frequencies for one fixed hardware configuration which generates losses at specific power transfer rates. The objective function of this optimization problem can be written generally as 

\begin{equation} \label{obj1}
f(L_r, L_m, C_r, k, f_i) = min \frac{1}{n}\sum^n_{i=1} (1-e_i) + \triangle P_i
\end{equation}

where $e_i$ denotes for the circuit's efficiency and  $\triangle P_i$ describes the deviation between the real output power $p_{r_i}$ and the target output power $p_{t_i}$  for all \textit{i} out of \textit{n} operation points. 

\begin{figure*}[ht]
\centerline{\includegraphics[width=0.95 \textwidth]{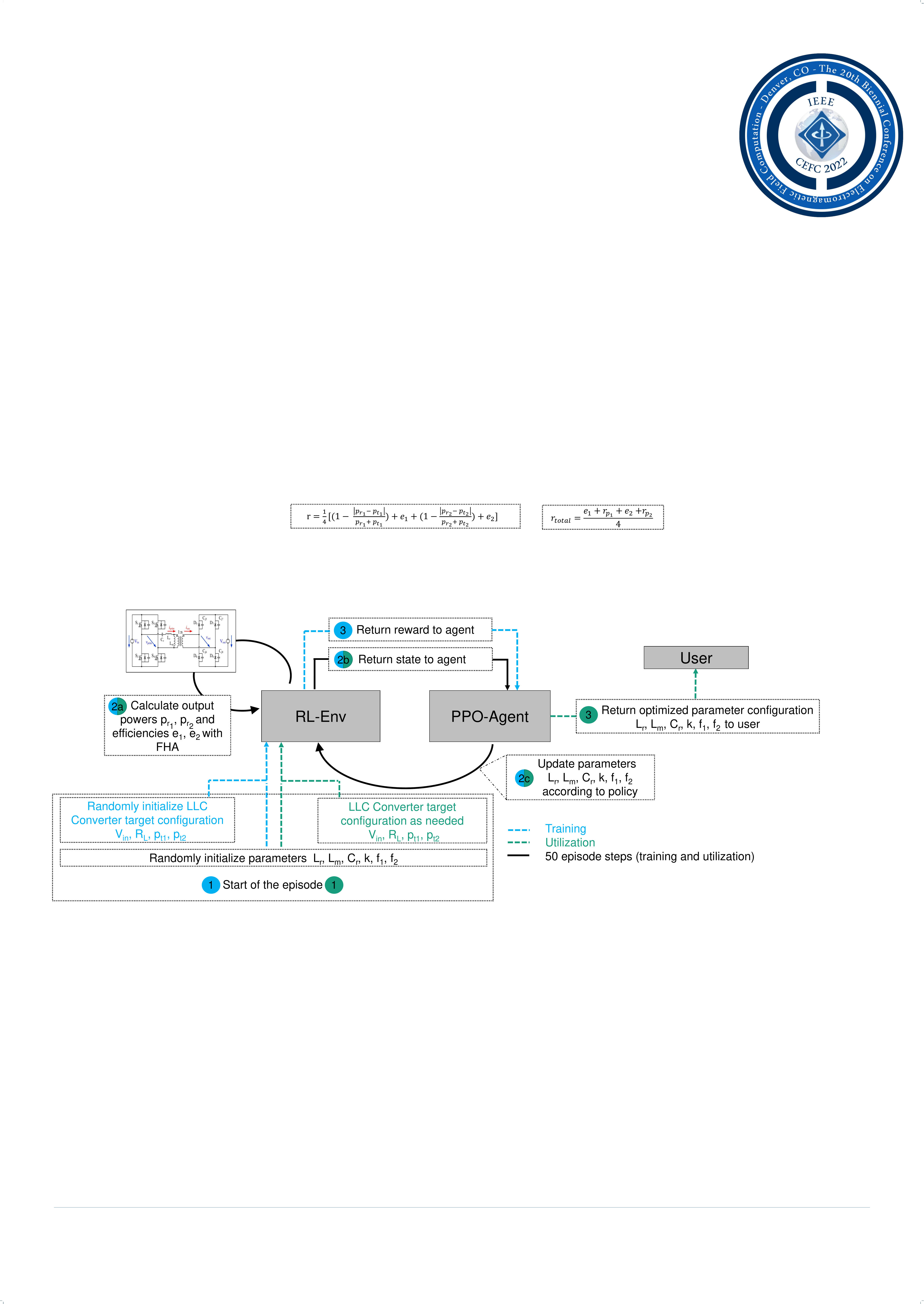}}
\caption{Training (blue+black): In the beginning of each episode, a LLC Converter target configuration is initialized randomly within the ranges of Tab. \ref{tab1} (1). Then the efficiencies and output powers are calculated for this parameter configuration by FHA (2a) and are fed back as part of the state to the PPO-agent (2b). The PPO-agent then updates the parameters according to its policy (2c) and this loop is repeated 50 times. At the end, a reward is calculated (Eq. \ref{reward}) and fed back to the agent. Utilization (green+black): The trained agent receives a LLC Converter target configuration specified by the user (1) and then optimizes the randomly initialized parameters for 50 steps (2a-2c). The final optimized parameter configuration can then be used by the user (3).}
\label{fig2}
\end{figure*}

\section{Reinforcement Learning}

In RL an agent interacts with an environment (see Fig. \ref{fig2}). At every step of interaction, the agent receives a state from the environment and takes an action depending on its policy. This policy is optimized through a reward signal which the agent receives after a predefined number of steps in the environment (also referred to as an episode). The goal of the agent is to converge towards a policy which maximizes the reward signal at the end of an episode in the given environment. 

\subsection{RL Agent}

In the field of RL, the Proximal Policy Optimization (PPO) algorithm belongs to the class of policy gradient model-free reinforcement learning agents using a special gradient clipping approach for the stabilization of the training process \cite{ppo}. Because of its training robustness and its capability of using continuous actions, we used PPO for our experiments. The implementation of the algorithm is taken from the RLlib Ray framework \cite{ray}.

\subsection{RL Environment}\label{RL}

The optimization problem of Section II is reformulated as a RL environment (OpenAI gym standard), where the goal of the agent is to update the LLC converter's fixed hardware configuration $L_r, L_m, C_r$ and $k$ as well as all frequencies $f_i$ in a predefined number of steps. An episode in this environment consists of  \textit{N} parameter update steps by the agent. For our experiments we use two operation points and 50 update steps for each episode (\textit{N}=50). 

\begin{table}[h]
\caption{Parameter ranges for LLC Converter}
\begin{center}
\begin{tabular}{|c|c|c|c|c|c|}
\hline
\rowcolor{light-gray}
$L_r$ [uH] & $L_m$ [uH] & $C_r$ [nF]& $k$ & $f_1$ [kHz] & $f_2$ [kHz] \\ [0.9ex] 
\hline
\rowcolor{light-gray}
0.1-100 & 0.1-100 & 1-1000 & 0.9-0.99 & 10-50 & 50-100 \\ [0.7ex]
\hline
\end{tabular}
\newline
\newline

\begin{tabular}{|c|c|c|c|}
\hline
$V_{in}$ [V] & $R_L$ [$\Omega$] & $p_{t_1}$ [W] &  $p_{t_2}$ [W]  \\[0.9ex]
\hline
400-500 & 30-40 & 100-300 & 4000-5000   \\ [0.7ex]
\hline
\end{tabular}
\label{tab1}
\end{center}
\end{table}

\paragraph{action shaping}

To map the output of the PPO to the necessary size for the parameter updates, the agents' actions, which lie in the ranges of [-1, 1] for each parameter, need to be scaled such that they fit the magnitude of the parameter ranges in Tab. \ref{tab1}. For each update step the maximal update step size is chosen to be 10\% of the parameter range of the respective parameter. Additionally, instead of linearly mapping the agents output to this step size by setting 1 $ \rightarrow $ 10\% and -1 $ \rightarrow $  -10\% we introduce a logarithmic scaling between 0 and +/- 10\%  to favor smaller step sizes by the agent.

\paragraph{reward shaping}

Generally, in RL the agent receives a reward from the environment at the end of each episode which is either 0 (for losing) or 1 (for winning). Since losing or winning is difficult to specify in the case of optimizing the parameters of a LLC Converter, we introduce a continuous reward between 0 and 1. As stated in Eq. \ref{obj1} the goal is to maximize the efficiency and minimize the output power deviation from our specified output power targets. Since the efficiency naturally scales between 0 and 1 it can be directly converted into a part of the reward. We therefore set the reward for the efficiency equal to the efficiencies themselves: $e_i \rightarrow r_{e_i}$.

To convert the output power deviation into a reward signal the canberra norm between the output powers and the target output powers are calculated. Since the goal is to minimize this deviation we set the output power reward to   

\begin{equation} \label{reward}
r_{p_i} = \Big(1 - \frac{|p_{r_i} - p_{t_i}|}{p_{r_i} + p_{t_i}}\Big).
\end{equation}

This output power reward $r_{p_i}$ is additionally scaled by introducing a scaling threshold $t_s$: If $r_{p_i}$ is smaller than $t_s$, then the total reward $r_{total}$ is set to 0, and the episode is considered as lost. If this criterion is met, then the $r_{p_i}$ is scaled between 0 and 1 according to $r_{p_i} = (r_{p_i} - t_s)\cdot(1/(1-t_s))$ and the total reward of the agent at the end of an episode for two operation points is calculated by

\begin{equation} \label{reward}
r_{total} = \frac{1}{4}\Big[r_{e_1} + r_{p_1} + r_{e_2} + r_{p_2}\Big].
\end{equation}

\subsection{Training}

The flow of an episode during training can be seen in Fig. \ref{fig2}. At the beginning of an episode, the parameters $L_r, L_m, C_r, k$ and $f_1$ and $f_2$  as well as the constant target values for the output powers (see Tab. \ref{tab1}) are sampled randomly within the predefined ranges. The output power and efficiency values for these values are computed by the LLC FHA simulation environment and fed back together with the target output powers and the current deviations and parameter values as input state to the agent. The parameters are then updated continuously by the agent for all $N$ steps according to its current policy. At the end of each episode, a reward is calculated according to Eq. \ref{reward} and fed back to the agent. The agent updates its policy in order to maximize this reward for each episode over the course of training and extracts a policy in the latent space of its neural network that enables a generic optimization of circuits for multiple and varying target output powers. The agent is trained for 80.000 episodes for varying target output powers and randomly sampled parameter configurations.

\subsection{RL based Optimization Strategy}

After the agent is trained it can be used to optimize specified LLC Converter configurations within one episode. The user can specify the configuration of interest as part of the initial state (see Fig. \ref{fig2}) and the trained agent optimizes for \textit{N} iterations the randomly initialized parameters for this configuration. The final parameter configuration can then be used by the user to start a detailed design process for the specified LLC Converter.

\section{Results}

\begin{figure}[ht]
\centerline{\includegraphics[width=0.49 \textwidth]{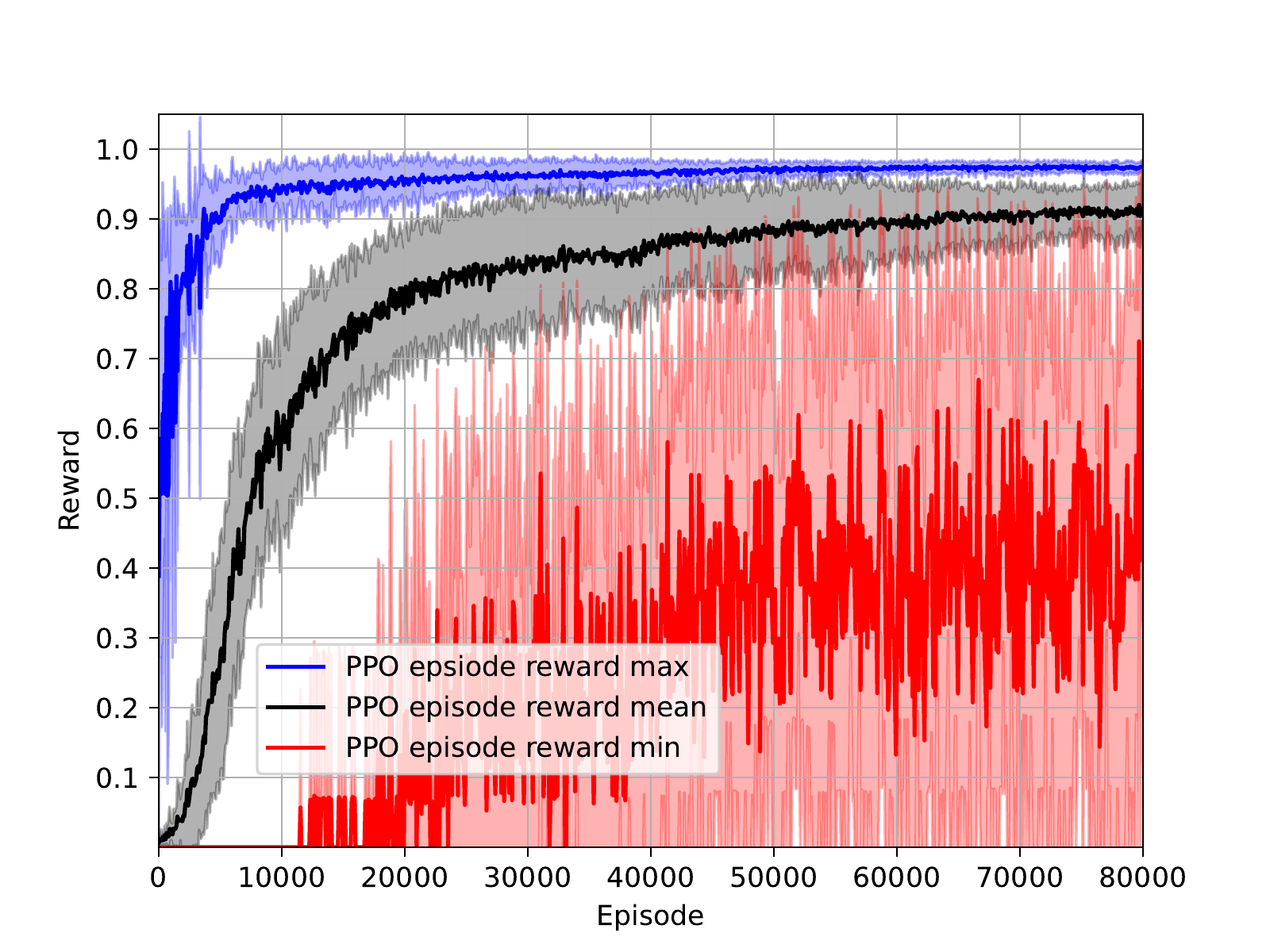}}
\caption{Training curves of 10 PPO-agents with different initializations of the weights of their neural networks. The mean of these trainings are the solid lines and the standard deviation are the shaded areas.}
\label{fig3}
\end{figure}

We trained 10 PPO-agents with two fully connected neural network layers of size 256 with different random weight initializations. We chose a learning rate of 1e-5 and a reward scaling threshold $t_s$ of 0.815. The mean (solid lines) and the standard deviation (shaded areas) of this training can be seen in Fig. \ref{fig3}. In the beginning of training, the agent quickly finds good configurations for some of the randomly initialized target LLC Converter configurations (blue line), but fails to do so for others (red line). On average the agent finds for the randomly sampled LLC configurations output powers and efficiencies corresponding to a reward of 0.91 (Eq. \ref{reward}).

\begin{figure}[ht]
\centerline{\includegraphics[width=0.49 \textwidth]{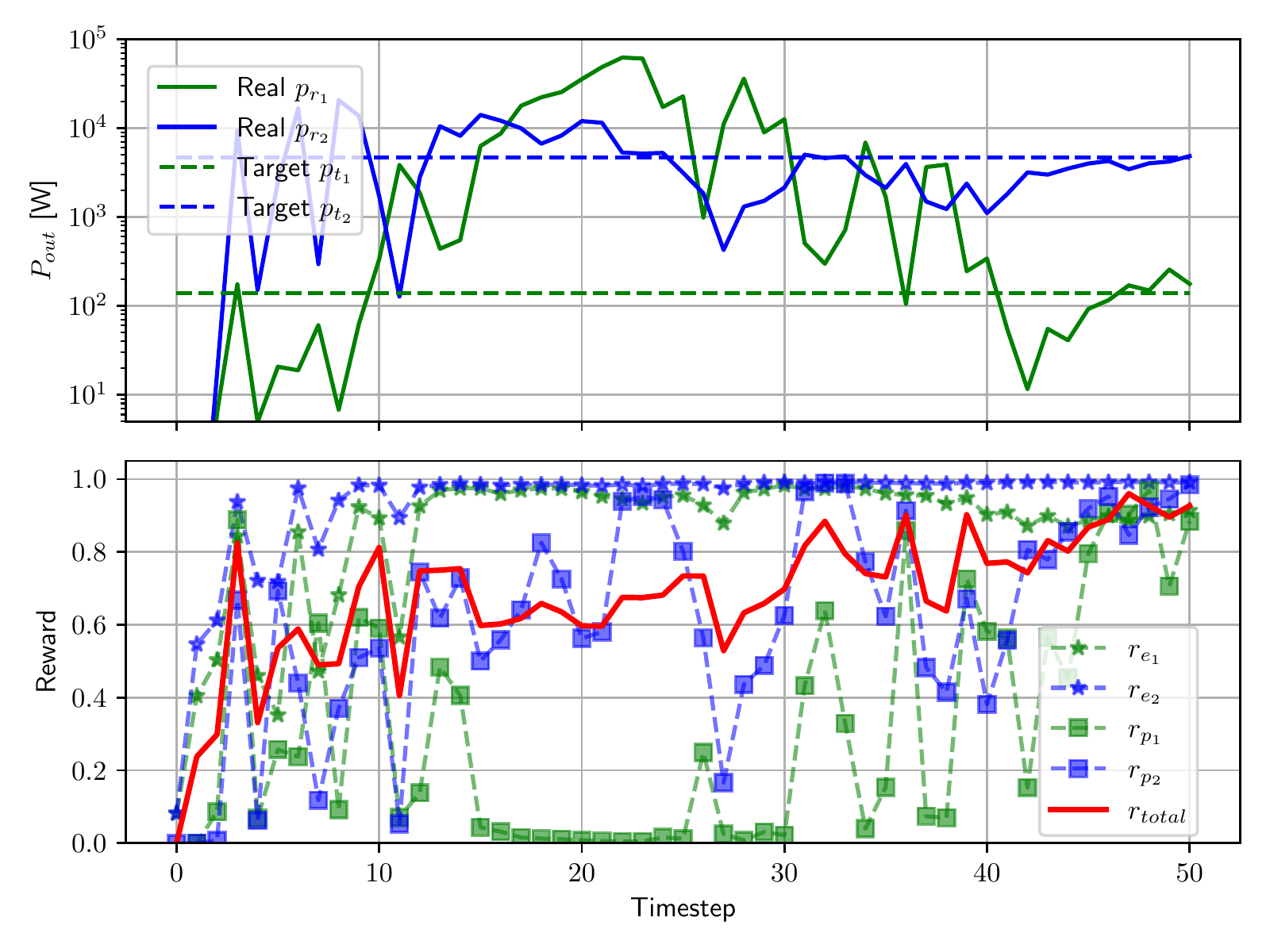}}
\caption{Exemplary episode of trained PPO agent after training: The output power values $p_{r_1}$ and $p_{r_2}$ approach the target power values $p_{t_1}$ and $p_{t_2}$ as the RL agent optimizes the parameters over the episode (upper). The corresponding reward is calculated as described in section \ref{RL} (lower).}
\label{fig4}
\end{figure}

Fig. \ref{fig4} shows an episode of a trained agent tackling an exemplary optimization problem with $p_{t_1}=140W$ and $p_{t_2}=4700W$. One can see in the upper plot, that the agent driven tuning process of manipulating the set of input parameters converge step-by-step to good solutions of the optimization problem. The two target operation points, marked by the dashed lines, are found within the scope of one episode (50 parameter update steps) with minor deviations. In the lower plot the correlation between the increasing values of the output power rewards, the efficiency rewards and the overall reward can be seen. The resulting parameter configuration has output power deviations from the specified target values of less than 5 \%  and efficiencies greater than 90\%. These optimized parameters can be used by the user to start a detailed design process.

\begin{figure}[ht]
\centerline{\includegraphics[width=0.46 \textwidth]{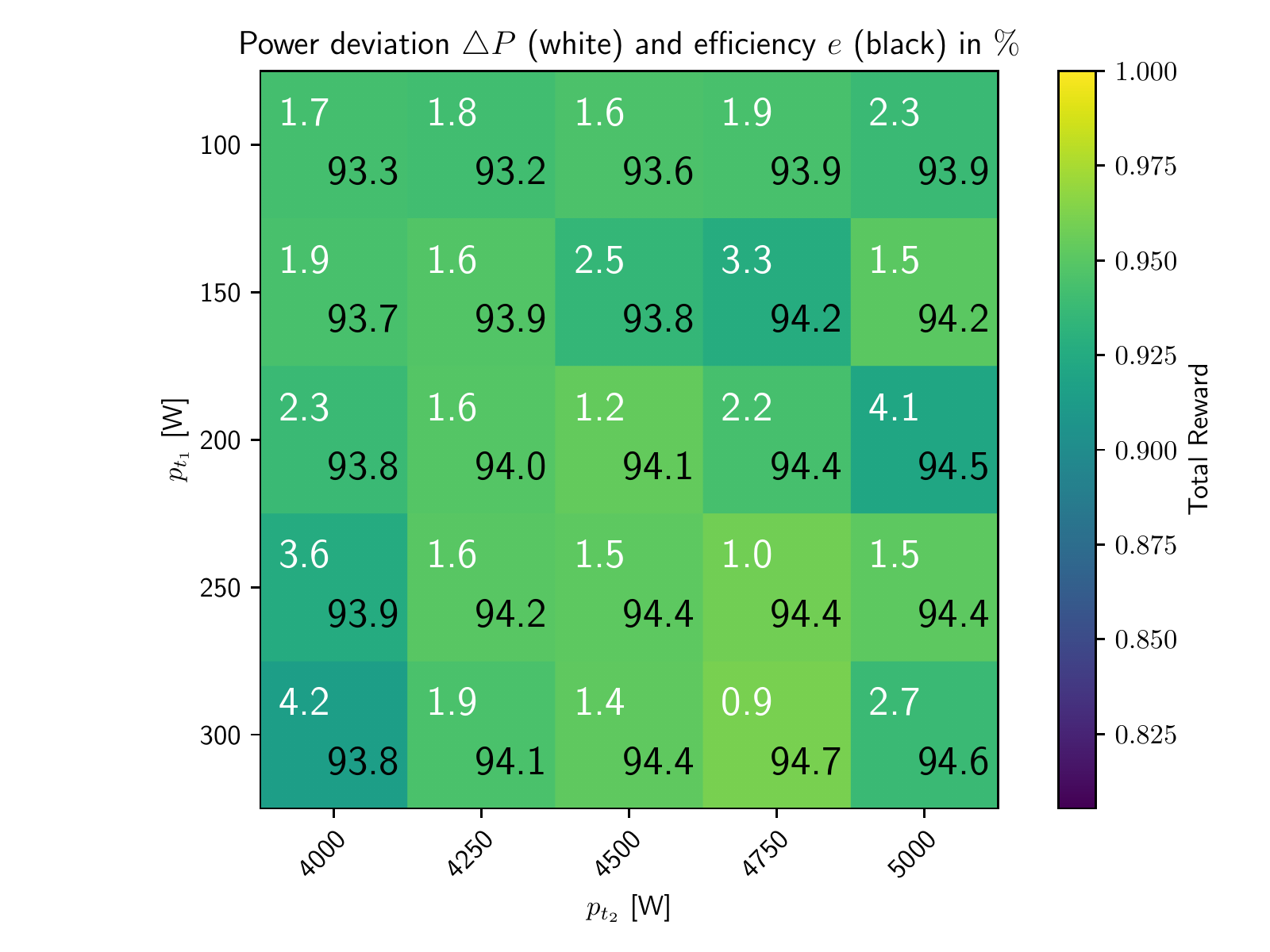}}

\caption{Testing matrix of trained agent: A grid search over the output powers $p_{t_1}$ and $p_{t_2}$ with the parameters ranges of Tab. \ref{tab1} is performed in a 5x5 matrix. We optimized each configuration with a trained agent with five different parameter initializations. The mean of the rewards, the output power deviations $\triangle P$ (white) and the efficiencies $e$ of the operation points (black) are plotted.}
\label{fig5}
\end{figure}

To further analyze the optimization capabilities of the agent, we performed a grid search of 25 different LLC Converter configurations (out of the ranges of Tab. \ref{tab1} bottom) for one of the 10 trained agents. For each of the 25 configurations we tested the agent for five different and randomly initialized starting parameters for $L_r, L_m, C_r, k$ and $f_1$ and $f_2$. We then averaged the performance over these five runs. In Fig. \ref{fig5} the matrix of this grid search can be seen. The average of the power deviation for the two operation points is plotted in white and has a maximum value of 4.2 \%. The efficiencies in black have a minimal value of 93.2 \%.

\begin{figure}[ht]
\centerline{\includegraphics[width=0.49 \textwidth]{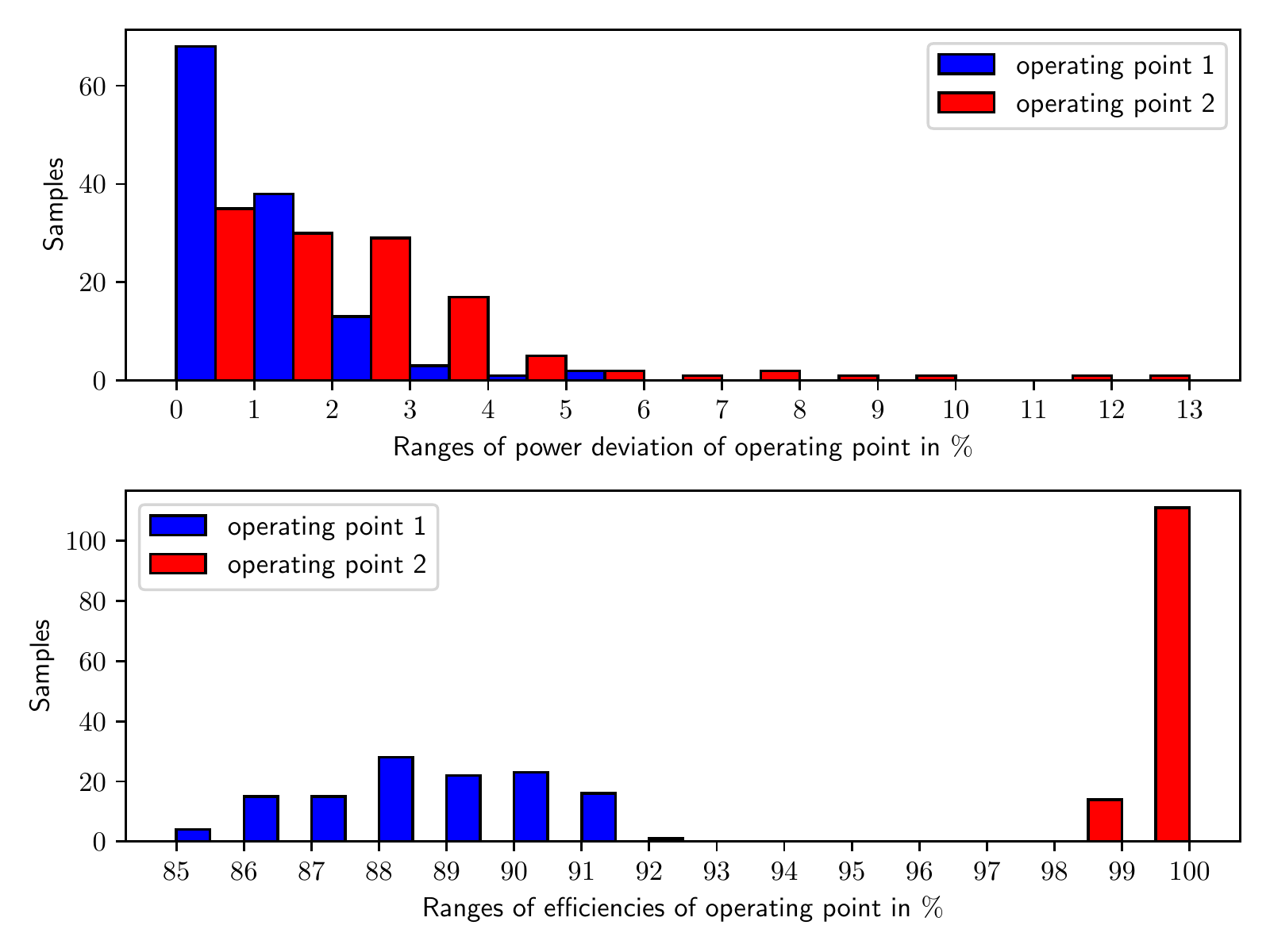}}
\caption{Distribution of output power deviation (upper) and efficencies (lower) in percent of all tested 125 samples for both operation points.}
\label{fig6}
\end{figure}

In Fig. \ref{fig6} all tested configurations are plotted: In the upper plot, the output power deviations and efficiencies in percent for the 125 total samples can be seen. For the first operation point (in the range from 100W to 300W), the maximal deviation lies between 5-6\% while the second operation point(in the range from 4kW to 5kW) shows deviations up to 12-13\%.

While the average efficiencies of the two operation points lie on average around 94\%, the lower plot of Fig. \ref{fig6} shows large differences between the two operation points: While all tested configurations for the second operation point show efficiency values greater 98\%, the efficiency values of the first operation point never reach values greater than 93\% and seem be primarily distributed around 89\%. 

\section{Conclusion and Outlook}

We showed that the parameters of the LLC converter can be optimized at multiple operation points by a pre-trained RL agent within 50 steps corresponding to a calculation time of less than a second. In contrast to existing GA strategies requiring several tens of thousands of evaluations for each optimization setup, the RL requires once four million evaluations enabling any optimization within the pre-defined optimization range within 50 steps. This provides great opportunities for solving computationally expensive optimization problems in engineering domains with similar, yet different requirements. The capability of such RL agents to solve similar optimization problems, but with different objective values and constraints, sets this approach apart from established optimization methods.  Future research will focus on wider parameter ranges, more complex LLC models and shorter training time of the agent, thus increasing its sample efficiency.

\section{Acknowledgments}

This work was supported by the Federal Ministry of Education and Research in the CODAPE "Kollaborative Entwicklungsumgebung für die Leistungselektronik" projekt via grant 16ME0356

\end{document}